\documentclass{article}

\usepackage[nonatbib]{neurips_2020}

\usepackage[utf8]{inputenc} 
\usepackage[T1]{fontenc}    
\usepackage{hyperref}       
\usepackage{url}            
\usepackage{booktabs}       
\usepackage{amsfonts}       
\usepackage{nicefrac}       
\usepackage{enumitem}
\usepackage{graphicx}
\usepackage{color}
\usepackage{amsmath}
\usepackage{multirow}
\usepackage[protrusion=true,expansion=true]{microtype}
\usepackage{floatrow}
\usepackage[table]{xcolor}
\definecolor{lightgray}{gray}{0.9}
\newfloatcommand{capbtabbox}{table}[][\FBwidth]

\title{Dialog Simulation with Realistic Variations for Training Goal-Oriented Conversational Systems}

\author{%
  David S.~Hippocampus\thanks{Use footnote for providing further information
    about author (webpage, alternative address)---\emph{not} for acknowledging
    funding agencies.} \\
  Department of Computer Science\\
  Cranberry-Lemon University\\
  Pittsburgh, PA 15213 \\
  \texttt{hippo@cs.cranberry-lemon.edu} \\
}

\begin{document}

\maketitle

\begin{abstract}
Goal-oriented dialog systems enable users to complete specific goals like requesting information about a movie or booking a ticket. Typically the dialog system pipeline contains multiple ML models, including natural language understanding, state tracking and action prediction (policy learning). These models are trained through a combination of supervised or reinforcement learning methods and therefore require collection of labeled domain specific datasets.
However, collecting annotated datasets with language and dialog-flow variations is expensive, time-consuming and scales poorly due to human involvement.
In this paper, we propose an approach for automatically creating a large corpus of annotated dialogs from a few thoroughly annotated sample dialogs and the dialog schema.
Our approach includes a novel goal-sampling technique for sampling plausible user goals and a dialog simulation technique that uses heuristic interplay between the user and the system (Alexa), where the user tries to achieve the sampled goal. We validate our approach by generating data and training three different downstream conversational ML models. We achieve $18-50\%$ relative accuracy improvements on a held-out test set compared to a baseline dialog generation approach that only samples natural language and entity value variations from existing catalogs but does not generate any novel dialog flow variations. We also qualitatively establish that the proposed approach is better than the baseline. Moreover, several different conversational experiences have been built using this method, which enables customers to have a wide variety of conversations with Alexa.
\end{abstract}

\vspace{-0.2cm}
\section{Introduction}
Goal-oriented dialog systems enable users to complete specific goals such as making restaurant reservations and buying flight tickets. User goals may often be complex and span multiple inter-dependent subgoals. This presents challenges for building accurate machine learning models that can understand user requirements provided over multiple turns, leverage knowledge sources, and learn to predict optimal actions for completing user goals with minimal friction. Building such models require thoroughly annotated dialog datasets with large dialog-flow and language variations. Unfortunately, only a few publicly available datasets~\cite{asri2017frames} meet such a standard and they cover only a limited number of domains. Wizard-of-Oz (WoZ) is a popular framework that can be used to collect additional human interactions that reflect the target user-dialog system interactions. While this setup does not require a working dialog system, a detailed knowledge of the domain, the desired system behavior and the annotation conventions are still necessary. These constraints impede building goal-oriented dialog chatbots across hundreds of domains in a relatively short time.

In this paper, we propose a dialog simulator that can generate thousands of dialogs given only a few annotated dialogs ("seed dialogs") and a dialog schema. In our experience, a novel domain will typically have at most 50 seed dialogs available. The schema is expected to include the list of Application Programming Interfaces ("APIs"), the catalogs of the entities, a few example utterances that users could use to interact with the system ("user-utterance templates") and the set of system response templates ("response-templates"). The generated dialogs include various natural dialog phenomena such as anaphora, entity sharing across multiple turns, users changing their mind during conversations and proactive recommendations. This approach is example-driven as it learns from a small number of provided dialog examples and doesn’t require encoding dialog flows as rigid rules.

The generated dialogs contain rich annotation using the domain-specific schema (APIs, response-templates, and entities) provided as part of the input. Hence they are suitable for training downstream supervised and reinforcement learning models for goal-oriented dialog applications. There are two main steps in the simulator: (1) Sampling a goal that user wants to achieve. We propose two different techniques to sample the goals: the \textit{Golden Goal Sampler} and the \textit{Markov Goal sampler} (Section~\ref{sec:user_goal}) (2) A user-system interplay where the user gradually reveals the goal to the system so that the system can fulfill the goal. Both user and system policy are heuristic based, where the user policy is motivated by the agenda-based user policy of~\cite{shah2018building} (Section~\ref{sec:interplay}). The proposed approach uses the user-system interplay to increase dialog flow variations, where Amazon MTurk is used to increase the language variations in the user utterances.

We evaluate the simulator in two different ways compared to a baseline. The baseline approach generates data with only language and entity variations sampled from existing catalogs, but no dialog-flow variations. First, we evaluate the quality of the simulated data via various novel qualitative metrics to establish that the simulated data using the proposed approach contains more variations than the baseline. Furthermore, we apply the simulated data to three different ML models in the context of goal-oriented dialog chatbots, i.e., Named Entity Recognition (NER), Action Prediction (i.e., predicting the API and system response templates) and Argument Filling (i.e., determining the arguments for an API or a response template).
We demonstrate that using the generated data with the proposed approach can improve the F1-score of NER by $18\%$ and the accuracy of Action Prediction and Action Signature by $21\%$ and $52\%$ relative respectively, over the baseline dialog generation approach. This validates the usefulness of our dialog simulation approach in generating diverse training data for training accurate downstream models.

\vspace{-0.2cm}
\section{Related Work}
\label{gen_inst}
Data requirements have been a primary bottleneck in training highly effective goal-oriented chatbots. There have been several prior efforts in collection of the annotated datasets. To address the time and cost requirements of WoZ setups, the authors in~\cite{shah2018building} proposed a Machines-Talking-To-Machines (M2M) framework, where a user and a system simulator interact to generate dialog outlines that are later transformed into natural language and expanded using crowd sourcing to create training data. There is also extensive prior work on user simulators that are used to interact with a dialog system to collect additional training data prior to deploying the system to real users~\cite{pietquin2005framework, cuayahuitl2005human, pietquin2006probabilistic}, including the use of entropy and other measures of dialog variation to evaluate conversational models (\cite{Pietquin2012}).

Our work extends the M2M framework in several directions. Instead of generating user goals randomly, we propose two different goal sampling techniques biased towards the goals observed in the seed dialogs in order to support variations of those dialogs robustly. In M2M, the system agent is geared towards database querying applications where the user browses a catalog, selects an item and completes a transaction. In contrast, our formulation does not require any knowledge of the purpose of each API but focuses instead on supporting a richer set of dialog patterns including complex goals, proactive recommendations and users correcting earlier provided entities. Furthermore, we propose a few intrinsic dialog metrics to evaluate the quality of the simulated data.

\vspace{-0.2cm}
\section{System Overview} \label{sec:overview}
Our proposed approach enables automatic generation of tens of thousands of dialogs that contain flow and language variations, and can be used for training conversational models for any goal-oriented dialog application. The application developer only needs to provide an order of ten annotated seed dialogs and the dialog schema (explained below).
To support domain-specific and cross domain dialog experiences, we follow the data-driven approach where we can provide seed dialogs covering the main uses cases we want to support. The dialog simulator learns dialog flows from the seed dialogs and generates novel synthetic goal-oriented dialogs.

\begin{figure}
\vspace{-0.2cm}
\begin{floatrow}
\hspace{-2.5em}
\ffigbox{%
  \centering \includegraphics[width=5cm,keepaspectratio]{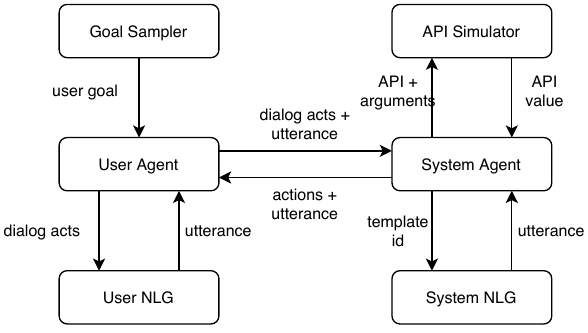}
}{%
  \caption{Simulator Architecture}\label{fig:architecture} %
}
\hspace{-2.5em}
\capbtabbox{%
\footnotesize
\centering
  \scalebox{0.73}
  {\begin{tabular}{p{0.05\textwidth}p{0.8\textwidth}}
U-1:& \texttt{What movie are playing in [Sunnyvale|c0] after [2 PM|t0]?} \\
S-2:&\texttt{call: FindMovies(location=\$c0,timeLowerBound=\$t0)} $\rightarrow$ \texttt{movies0} \\
S-3:& \texttt{nlg: Tenet is playing in AMC Theater at 4 PM} \\
U-4:& \texttt{tell me more about the [4 PM|t1] show of [Tenet|m0]} \\
S-5:& \texttt{call: SelectShow(time=\$t1, movieName=\$m0)} $\rightarrow$ \texttt{show0} \\
S-6: & \texttt{nlg: OK. The available ticket type is adult ticket} \\
U-7: & \texttt{Book [two|c0] [adult|type0] tickets for this show} \\
S-8: & \texttt{BookTickets(show=\$show0, count=\$c0, type=\$type0) -> booking0} \\
U-9:& \texttt{Ok thank you}\\
S-10: & \texttt{nlg: Thank you for using Atom Tickets}
  \end{tabular}
  }
}{%
  \caption{A dialog with markup language annotations.} \label{table:acolade_example}%
}
\end{floatrow}
\end{figure}

The simulator is structured in two distinct agents that interact turn-by-turn: the user and the system. Figure~\ref{fig:architecture} shows the overview of how each simulator component communicates to each other. User agent samples a fixed user goal at the beginning of the conversation. Agents communicate at the semantic level through dialog acts. Having the exact information associated with each turn allows us to define a simple heuristic system policy, whose output can be used for supervised training labels to bootstrap models. Note that the user policy is also heuristic-based. In each conversation, user agent gradually reveals its goal and system agent fulfills it by calling the APIs.
The system agent simulates the API call by randomly sampling a return value without actually calling the API and chooses an appropriate response action. Depending on the value returned by the API, the chosen response is associated with dialog acts. The system agent gradually constructs the estimated user goal and makes proactive offers based on the estimated goal. The interplay loop is described in Section~\ref{sec:interplay}.
The dialog acts generated through interplay is also used to interface between agents and their template-based Natural Language Generation (NLG) model. After sampling the dialog acts from their policy, each agent samples the surface-form from available templates corresponding to the dialog acts.
In addition to enriching the dialog flows, we use crowd-sourcing though Amazon Mechanical Turk (MTurk) to enrich the natural language variations of the user utterance templates.
Goal sampling and the interplay loop provides dialog flow variations while crowd-sourcing enriches natural language variations, both of which are required for training robust conversational models.

\vspace{-0.2cm}
The inputs to simulator include the domain-specific APIs, entity types with catalogs, system response templates, user-utterance templates, and seed dialogs.  Each API is defined with some arguments annotated with entity types and a response NLG template. API arguments can be required or optional. System response templates also have input arguments and contain different language templates annotated with different dialog acts. To enrich the language variations, the application developer can provide catalogs for the entity types. E.g., fantasy, drama, and action can be the catalog values for the entity \textit{Genre}. Although our system auto-generates user-side utterance templates from the language of the annotated seed dialogs, the developer can additionally provide user utterance templates. Finally, the seed dialogs serve as training examples, and are annotated in a markup language, as shown in Table~\ref{table:acolade_example}. The dialog simulator generates output dialogs annotated in the same markup format.

\vspace{-0.2cm}
\section{User Goal}
In order to train a robust task-oriented neural dialog system, it is important to simulate diverse but consistent dialogs. This section first introduces a concept of user goal, which will serve as backbone for a dialog and ensure consistency. Then, we discuss how to obtain diversity of dialog flows by generating diverse user goals.

\subsection{User Goal Representation}  \label{sec:user_goal}

To reconcile the behavior of the users of a task domain with the desired system actions, we assume that the APIs of a task domain have been designed to support a set of \textit{user intents}. By definition, a user intent can be communicated by the user in a single utterance but it can be fulfilled by the system by calling one fixed API or a fixed group of APIs. For example, consider a user that wants to browse available movies in Sunnyvale after 2 PM and communicates that intent in user utterance U-1 of seed dialog in Table~\ref{table:acolade_example}.

The system may fulfill that request by calling an API with a set of argument values (e.g. \texttt{location} and \texttt{time}) inferred from the user utterance:

\vspace{-0.2cm}
\begin{verbatim}
FindMovies(location="Sunnyvale", time="2 PM") -> movies0
\end{verbatim}

The API call produces return values (e.g. \texttt{movies0}) that become part of the dialog context and may be used as arguments to later API calls if required by the dialog continuation.

In practice, a single dialog may involve several inter-dependent intents, all referring to a common set of entities elicited from user utterances throughout the dialg. For example, assuming the movie selection \texttt{movies0} suits the user, they may then continue the dialog by asking for more details about a specific showtime or book some previously selected tickets (in U-4 and U-7 of Table~\ref{table:acolade_example}, respectively), which can be fulfilled by executing the actions:

\vspace{-0.2cm}
\begin{verbatim}
SelectShow(time="4 PM", name="Tenet", movies=$movies0) -> show0
...
BookTickets(show=$show0, count="2", type="adult") -> booking0
\end{verbatim}

As this example dialog illustrated, some API arguments values are filled by user-provided values while others are filled by API return values (from the same compound intent or from previous ones). At the start of a dialog, we assume that a user has in mind a sequence of intents to fulfill, a set of entity values (e.g. "Sunnyvale", "2 PM") and knows how to connect these values (e.g. fill \texttt{location="Sunnyvale"}) and those that will be returned by intents (e.g. \texttt{show=\$show0}). A \textit{user goal} is defined as such a sequence of interconnected intents together with user-provided entity values. This fixed goal ensures that the user behaves in a consistent goal-oriented manner and may also serve as a basis to define a goal completion reward signal at the end of the dialog.

\subsection{Generating User Goals}

A small set of representative user goals is extracted from the provided seed dialogs by converting each annotated API call graphs into a corresponding user goal. The \textit{Golden Goal Sampler} randomly samples goals from this set with replacement, providing a strong bias towards the observed seed dialogs so that they are well supported by the conversational models. The user-provided entity values will be resampled in each golden goal sampling.

To provide additional diversity, we also introduce the \textit{Markov Goal Sampler} which samples user goals from a generative sequence model
\begin{equation} \label{eq:1}
P(goal) = \prod_i P(intent_i|(intent_j)_{j<i})
\end{equation}
This goal model is estimated from the provided
seed dialogs, biasing towards the intent transitions and entity
sharing patterns present in those dialogs. However, we also generate novel
goals that were not observed in the seed dialogs, which enable our
conversational models to generalize. We ensure that all generated goals are
valid, meaning they are consistent with the developer schema. For example, we
ensure APIs have all of their required arguments filled in, entity sharing only
happens across API arguments of the same entity type. Note that for simplicity,
our generative goal model makes the assumption that the $i^{th}$ intent is
only dependent on the $(i-1)^{th}$ intent (i.e.\ a linear Markov chain assumption)
in the current implementation.

This formulation naturally extends to many domains, by generating goals and
corresponding dialogs that contain APIs spanning multiple domains. For the
multi-domain use case, we assume that entities can be shared across domains if
and only if they follow a built-in common schema (e.g., \textit{Time},
\textit{Date}, \textit{Address}) and thus can be understood by all domains. As a result, the
simulation process can construct goals where built-in entities that are
mentioned by the user or returned by an API in one task domain can be shared with the
APIs of other domains that take such entities as input arguments.

\section{Dialog generation through User-System Interplay} \label{sec:interplay}
The simulator consists of two agents: user and system. As discussed in Section~\ref{sec:user_goal}, at the start of each simulated conversation, the user
agent samples a fixed goal. The agents then interact in a interplay loop where the user gradually reveals their goal for the system to cater.

The agents communicate at the semantic level through sequences of dialog acts as presented in Section~\ref{sec:dialog-acts}.
Building this dialog-act-level meaning representation of the dialog state at each turn allows us to impose heuristic (deterministic) or neural policies for the user and the system that act on this dialog act meaning representation.
Using heuristics similar to the agenda-based policy of \cite{schatzmann2007agenda}, the user selects a sequence of dialog acts at each turn. Those acts are then passed to the system agent together with a consistent turn utterance obtained from the user NLG subsystem. Depending on the dialog context and associated past and current dialog acts, the policy of the system agent may select appropriate APIs, argument values, and format of the response utterance.

In the case of a complex goal with multiple inter-dependent intents, the user agent will iteratively serve each successive intent and, for each, enter a sub-dialog like Table~\ref{table:acolade_example}. Since API return values are not known in advance and may be non-deterministic, it is possible that the system is unable to fulfill an intent. In those cases, the simulated user may abandon the intent, remove from their goal any dependent intent and, if possible, carry on with the conversation.

In the remainder of this section, we highlight several example heuristics implemented in the simulated user and system policies to inject dialog flow variation into the generated dialogs. Those flows allow our agents and the neural system policy to support general dialog phenomena beyond the strict patterns that appear in the input schema described in Section~\ref{sec:overview}.

\subsection{Self-play Communication with Dialog Acts}
\label{sec:dialog-acts}

During the course of a dialog simulated through the interplay mechanism described in Section~\ref{sec:interplay}, the user and system communicate to achieve the user’s goal. They do so by exchanging dialog acts: a delexicalized grammar encoding information about the structure of a general task-oriented dialog. Using dialog acts to encode task-related dialog information reduces the dimensionality of the action space for heuristic or learned policies to operate, while preserving information about intents and the structure of dialog that generalizes across users and task domains.

\begin{table}[t]
\footnotesize
  \caption{Dialog Acts and Usage}
  \label{dialog_acts}
  \centering
  \resizebox{\columnwidth}{!}{%
    \begin{tabular}{lll}
      User Dialog Acts & &  \\
      \toprule
      inform(intent) &  U informs S of an intent  & "please find me a movie showing"   \\
      inform(entity) &  U informs S the value of an intent argument  & "for Tenet"   \\
      affirm(intent) &  U affirms that an intent is correct  & "yes, I want to find a movie showing"   \\
      affirm(entity) &  U affirms that an intent argument is correct  & "yes, for Tenet"   \\
      deny(intent) &  U denies that an intent is correct  & "no, I don't want to try again"   \\
      deny(entity) &  U denies that an intent argument is correct  & "no, not for Tenet"   \\
      \bottomrule
      & & \\
      System Dialog Acts & &  \\
      \toprule
      inform(entity) &  S informs U the value of an intent argument & "the movie you requested is showing nearby at 4 PM"   \\
      confirm(intent) &  S confirms that an intent is understood  & "are you ready to book the ticket?"   \\
      confirm(entity) &  S confirms that an intent argument is understood  & "for Tenet at 4 PM"   \\
      offer(intent) &  S proposes an intent to U  & "would you like to find an movie showing?"   \\
      offer(entity) &  S proposes an intent argument to U  & "how about Tenet?"   \\
      \bottomrule
    \end{tabular}
  }
\end{table}

Table~\ref{dialog_acts} provides the definitions of the primary dialog acts used in simulation, as well as example natural language invocations of those dialog acts. Several other types of dialog acts exist (e.g. failure(intent), bye(), and repeat(), etc.) and encode other conditioning information useful for dialog policy decision making. Given this dialog grammar in terms of dialog acts, it is also useful to think about sequences of dialog acts - paired with dialog act arguments - as representing dialog “super structures” such as offers, confirmations, and corrections.
\begin{verbatim}
S: {"would you like to book Tenet at 4 PM"}
-> "offer(BookTickets),offer(movieTitle),offer(showTime)"
U: {"no thank you, I would like to book it at 17:00"}
-> affirm(BookTickets),affirm(movieTitle),deny(showTime),inform(showTime)"
\end{verbatim}

The distribution over these dialog act sequences represents a measure of the diversity of a set of dialogs in terms of the degree of variation in the semantic content of the dialog act-argument pairs found in those dialogs. For the full sequence containing all dialog acts appearing in a particular dialog, measures of sequence variation correlate with the diversity of the underlying synthetic dialog dataset, properties studied in Section~\ref{sec:simulator-metrics}.

\subsection{Additional Dialog Variations}

{\bf User Correction:} In goal-oriented conversations, users often change their mind during the course of the conversation. For example, while booking a movie ticket user may decide to purchase three adult tickets but could eventually change their mind to book only two tickets; while booking a rideshare ride, users may decide to change the pick-up address in the middle of the interaction. This behavior (correcting a previously mentioned api argument) is general in the sense that it can appear in many different task domains, but may be difficult to extrapolate from a small number of input dialogs. Unless we generate these behavior in the training data, the conversational models will not respond flexibly in these situations. Hence, we propose a heuristic approach to simulate "correction" behavior, where users change their mind and the system responds accordingly, in a general, non-task-specific context.

\begin{enumerate}[noitemsep, leftmargin=*, topsep=0pt]
\item Given the goal representation (Section~\ref{sec:user_goal}) of the user's intent, alternative goals are sampled that contain the same API call structure, but alternative argument values that the user might provide.

\item If user corrects the API argument before the corresponding API is called, we update the user and system states with the new value.

\item If user corrects an API argument \emph{after} the corresponding API is called, the system recalls that API with the updated value and returns the updated information to the user.

\item During simulation, the user may randomly change their mind in any turn about an earlier informed entity.

\end{enumerate}

{\bf Proactive System Offers:} Another important non-task-specific conversational behavior is the system's ability to suggest an appropriate next action
based on the conversation history, without requiring invocation by a specific user utterance. Enabling proactive offers in the system policy facilitates exploration of the available API functionality in a manner consistent with human conversation. In a multi-domain system setting, the ability to easily explore the set of valid system interactions is especially important. To implement system policies with proactive offer capabilities, we estimate at each turn a distribution over the
next user actions/sub-intents and sample the next API to offer based on this distribution. The algorithm is calibrated to ensure relevance and variety for proactive API offers in practice.

\vspace{-0.2cm}
\section{Evaluation}

In this section, we measure the utility of the proposed dialog simulator in two ways. First we measure the amount of variation under the dialog act grammar defined in Section~\ref{sec:dialog-acts} that the simulator introduces through goal sampling and interplay. Then, performance is shown to improve by introducing this synthetic dialog variation to the training data for the downstream conversational models.

\vspace{-0.2cm}
\subsection{Variation in Synthetic Dialogs}
\label{sec:simulator-metrics}
To measure the amount of dialog variation present in a sample of dialogs represented in the format of Table~\ref{table:acolade_example}, we construct three estimators for moments of the distribution of dialogs, where dialogs are represented as deserialized sequences of dialog act-argument pairs.
\begin{itemize}[noitemsep, leftmargin=*, topsep=0pt]
  \item Number of "turns" per dialog, i.e. the number of utterances (by both user and system) present in the dialog
  \item Number of unique, complete-dialog dialog act sequences that are present in the dataset
  \item Entropy \footnote{Entropy of dialog variation to evaluate generative dialog models has also been used in \cite{Purgai2019} } of the distribution of complete-dialog dialog act sequences
  \\ $ \hat{\mathcal{I}}(\hat{p} ) = -  \sum_{i=1}^n \hat{p}(x_i) \log( \hat{p}(x_i))$ with $\hat{p}$: empirical distribution of dialog act sequences
\end{itemize}

We estimated these quantities in distribution by generating $10,000$ dialogs for five different task domains, each represented by a collection of dialogs written in the input format of Table~\ref{table:acolade_example} (i.e. seed dialogs). Each set of seed dialogs contains different numbers of intents (APIs), entities (API arguments) and different complexity of system \& user response interfaces. This design allows us to measure the amount of dialog variation introduced by the simulator to the raw seed dialogs for different structures of conversational system designed for different tasks as presented in Table~\ref{table:simulator-metrics}.

\begin{table}[th]
\footnotesize
\vspace{-0.1cm}
  \caption{Measures of Dialog Variation for Dialogs of Different Task Domains}
  \label{table:simulator-metrics}
  \centering
  \rowcolors{1}{}{lightgray}
  \begin{tabular}{|c|p{0.7cm}|p{1.1cm}|p{1cm}|l|l|l|p{1.5cm}|p{1.55cm}|p{1cm}|}
    \hline
    \multirow{2}{*}{Task} & \multirow{2}={\# of Intent} & \multirow{2}={\# of Entities} & \multirow{2}={Sampler Type} & \multicolumn{3}{|c|}{Turn per Dialog} & \multicolumn{3}{|c|}{Dialog Act Sequence\protect\footnotemark } \\
    \cline{5-10}
    & & & & Mean & P-75 & P-95 & \# of Unique Seq. & Fraction of Unique Seq. & Entropy\protect\footnotemark \\
    \hline
    & & & Base & 10.3 & 12 & 14 & 24 & 1.3\% & 3.16 \\
    \cline{4-10}
    & & & Golden & 9.4 & 12 & 16 & 428 & 11.9\% & 5.03 \\
    \cline{4-10}
    \multirow{-3}{*}{task1} & \multirow{-3}{*}{11} & \multirow{-3}{*}{25} & Markov & 8.3 & 10 & 16 & 2013 & 55.9\% & 6.95 \\
    \hline
    \multirow{3}{*}{task2} & \multirow{3}{*}{4} & \multirow{3}{*}{8} & Base & 6.1 & 6 & 10 & 11 & 0.6\% & 2.40 \\
    \cline{4-10}
    & & & Golden & 10.0 & 12 & 14 & 654 & 18.2\% & 4.89 \\
    \cline{4-10}
    & & & Markov & 8.1 & 10 & 14 & 1167 & 32.4\% & 5.66 \\
    \hline
    & & & Base & 7.5 & 8 & 10 & 9 & 0.5\% & 2.16 \\
    \cline{4-10}
    & & & Golden & 10.4 & 12 & 16 & 1929 & 53.6\% & 6.86 \\
    \cline{4-10}
    \multirow{-3}{*}{task3} & \multirow{-3}{*}{4} & \multirow{-3}{*}{12} & Markov & 11.1 & 14 & 18 & 2216 & 61.6\% & 7.13 \\
    \hline
    \multirow{3}{*}{task4} & \multirow{3}{*}{2} & \multirow{3}{*}{4} & Base & 5.0 & 6 & 6 & 2 & 10.0\% & 0.69 \\
    \cline{4-10}
    & & & Golden & 9.2 & 10 & 14 & 24 & 60.0\% & 2.90 \\
    \cline{4-10}
    & & & Markov & 10.6 & 11 & 17 & 30 & 75.0\% & 3.22 \\
    \hline
    & & & Base & 12.3 & 16 & 24 & 24 & 1.3\% & 3.05 \\
    \cline{4-10}
    & & & Golden & 15.6 & 18 & 32 & 1395 & 38.8\% & 6.19 \\
    \cline{4-10}
    \multirow{-3}{*}{task5} & \multirow{-3}{*}{6} & \multirow{-3}{*}{6} & Markov & 13.1 & 16 & 30 & 1972 & 54.8\% & 6.53 \\
    \hline
\end{tabular}
\end{table}
\addtocounter{footnote}{-1}
\footnotetext{Dialog act sequenes are defined as comma-separated strings of (dialog act name, dialog act argument) pairs and are equivalent if and only if their string representations match exactly}
\addtocounter{footnote}{1}
\footnotetext{Entropy is defined as the in-sample estimator of $p$'s informational content $-E[log(p(seq))]$, where $p()$ is the empirical distribution over dialog act sequences}
\addtocounter{footnote}{1}

The baseline, Base sampler, approach simply resamples dialogs that are identical in logical structure to the seed dialogs. It adds no dialog act level variation to the generated dialogs, only language variation in via catalog and template sampling. Base sampler dialogs therefore  exhibit very little variation entropy.
The \textit{Golden Goal Sampler} introduces a structured representation of user intent resulting in longer dialogs on average due a thicker tail of longer and more complex dialog samples.
The \textit{Markov Goal Sampler} introduces more nuanced variation by sampling different configurations of \emph{intents \& arguments} resulting in significant increase in the diversity of dialog act sequences represented in the sample, and in the entropy of the distribution of dialog act sequences.

Entropy increases successively (from base to golden to markov) with increase in conditioning information available to sampler, as resulting dialogs become more varied and finer with respect to amount of semantic information encoded in the dialog act grammar.

\subsection{Conversational System Performance with Synthetic Dialog Generation}

We use data generated with the proposed dialog simulator to train three models: a Named Entity Recognition (NER) model that tags entities in the user utterance, an Action Prediction (AP) model that predicts which API or system response should be called next, and an Argument Filling (AF) model that fills the (possibly optional) action arguments with entities available in context. The latter two models represent the “policy” of the next system action.

For the NER task, we use a bi-LSTM model extending \cite{Ma2016}. To incorporate information from dialog history we extract turn- and dialog-level token sequences from the context, pass them through context encoders, and concatenate the feature representations to obtain the final dialog representation. Additionally, we incorporate domain-specific catalog-based features for entity values similar to \cite{Williams2019}. For the AP task, we pass the dialog context features enhanced with output from the NER model through a feedforward layer to output a distribution over all actions within the domain. We utilize n-best action hypotheses to improve error robustness of the AP step. For the AF task, we model the problem as a variation of neural reading comprehension \cite{Chen2018} and adapt the model architecture proposed in \cite{Gao2019}. We impose constraints on the decoder to only fill arguments with entities of the correct type according to the action schema (e.g. the set of prespecified application-specific API signatures).

To assess the quality of the policies learned by the AP \& AF models, we evaluate the composite model output against held-out test sets collected and professionally annotated with ground truth NER tags and API/system response signatures through a Wizard-of-Oz paradigm for a ticket booking domain. The test set consists of $50$ dialogs with an average length of $7.7$ turns. We measure the F1 scores for spans \cite{sang2003introduction} of entities to evaluate NER performance, as well as AP and “action signature prediction” (ASP) accuracy to quantify the performance of the system policy. An ASP is counted as correct when both the action and all the corresponding arguments are predicted correctly; this measure proxies for the turn-level accuracy experienced by a user interacting with the system agent.

Our experimental design is as follows. For three configurations of the dialog generation methods described in Section \ref{sec:simulator-metrics}, we generate $10,000$ dialogs, train NER, AP, \& AF models, and evaluation those models on the test set. We repeat this procedure $5$ times for each configuration and the average performance over the trials is reported in Table~\ref{table:ticketbot-results}. The results reflect the impact on the downstream tasks of training with dialog data synthesized via our proposed generation method. Noted that both AP and ASP accuracy are evaluated given the ground truth of the NER results.

In the first configuration (C1), we use the Base sampler to resample dialogs. In the second configuration (C2), we generate dialogs using the \textit{Golden Goal Sampler} through user-system interplay, which introduces logical variation in the dialog act sequence. In the third configuration (C3), we generate dialogs using both the Golden ($40\%$ of dialogs) and Markov ($60\%$ of dialogs) Goal Samplers through user-system interplay, which introduces logical variation in both entities and intents.

\begin{table}[th]
\footnotesize
\vspace{-0.1cm}
  \caption{NER span F1-score, AP accuracy and ASP accuracy on the ticket booking test set, averaged over $5$ runs.}
  \label{table:ticketbot-results}
  \centering
  \begin{tabular}{lccc}
    \toprule
    \multicolumn{1}{c}{\textbf{Goal Sampler}} & \multicolumn{1}{c}{\textbf{NER Span}} & \multicolumn{1}{c}{\textbf{Action Prediction }} & \multicolumn{1}{c}{\textbf{Action Signature }}   \\
      & \multicolumn{1}{c}{\textbf{Relative F1}} & \multicolumn{1}{c}{\textbf{Relative Accuracy}} & \multicolumn{1}{c}{\textbf{Relative Accuracy}}   \\
    \midrule
    Golden (C2)            & $+17.22\%$ & $+18.90\%$ & $+49.17\%$ \\
    Markov and Golden (C3) & $+18.50\%$ & $+20.92\%$ & $+52.80\%$ \\
    \bottomrule
  \end{tabular}
\end{table}

We observe that models trained on data generated with C2 and C3 significantly outperform the models trained on data generated with C1. In case of NER, we see a $17-19\%$ improvement over the baseline; for AP accuracy, the improvement is $18-21\%$. For ASP accuracy, which is most reflective of the quality of the system policy experienced by a user, introducing logical variation to the training data via C2 and C3 improves the resulting model performance on the test set by $49-53\%$.

We hypothesize that these significant improvements in the quality of system policies learned from data simulated by our proposed methodology are derived primarily from the elevated amount of dialog variation present in the C2 and C3 training sets relative to the baseline seed dialogs. The seed dialogs are few in number and are not able to fully encode the range of conversational flows that a potential user might traverse in achieving her goal – but when augmented with the Golden and Markov samplers, the training data is able to better reflect these alternative variations appearing in the test data or in real user interactions.

\begin{figure}[!htb]
\minipage{0.32\textwidth}
  \includegraphics[width=\linewidth]{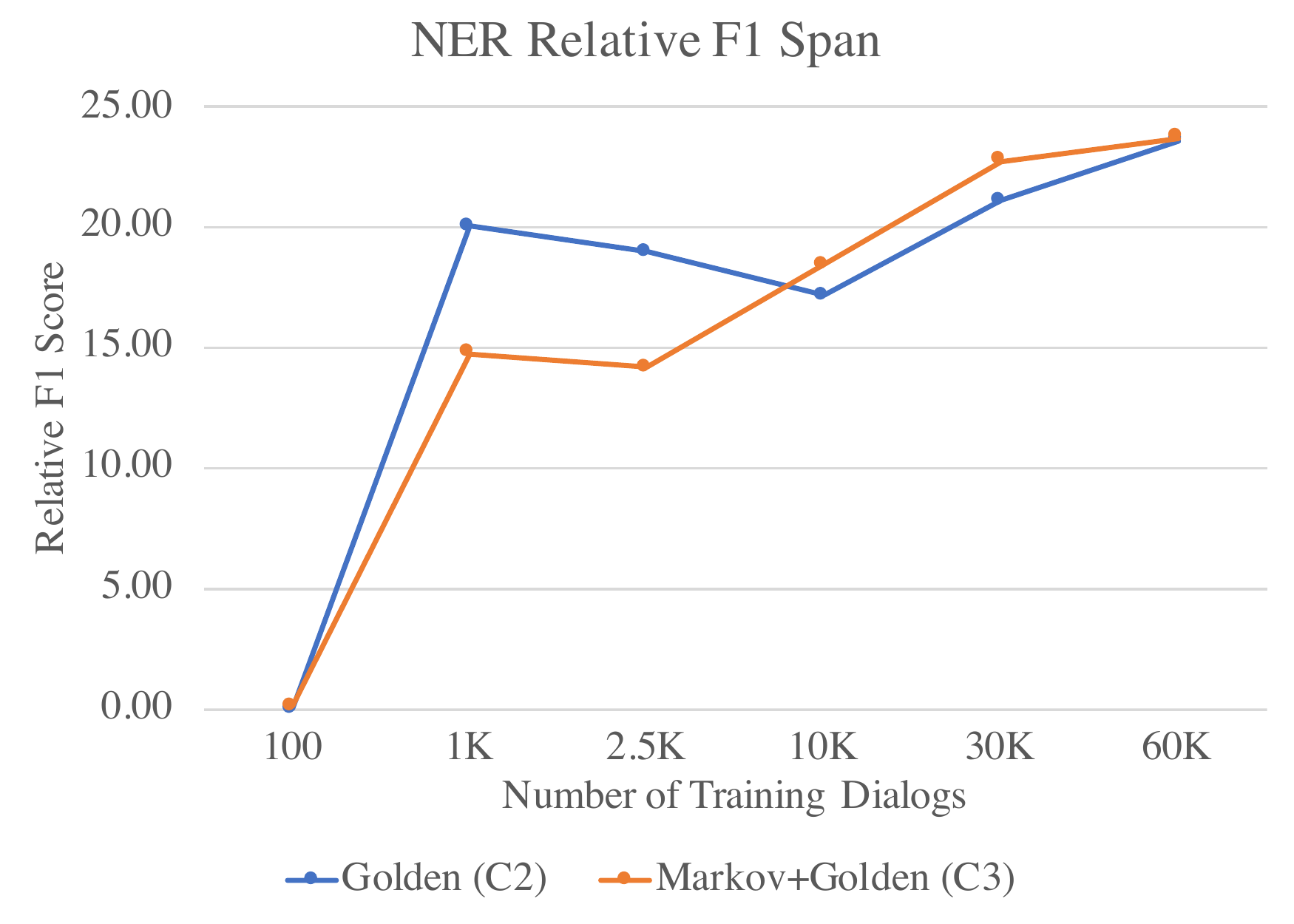}
\endminipage\hfill
\minipage{0.32\textwidth}
  \includegraphics[width=\linewidth]{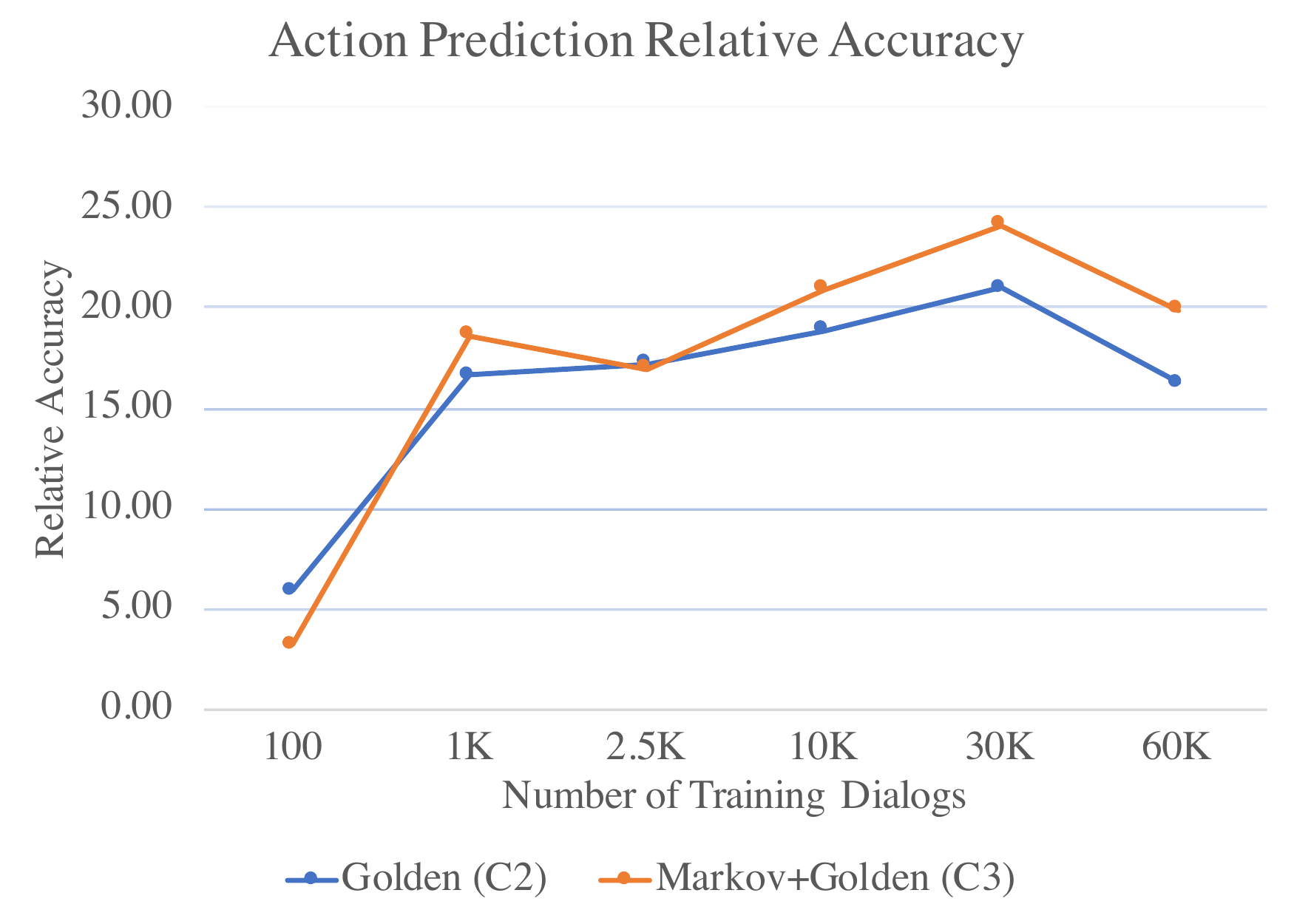}
\endminipage\hfill
\minipage{0.32\textwidth}
  \includegraphics[width=\linewidth]{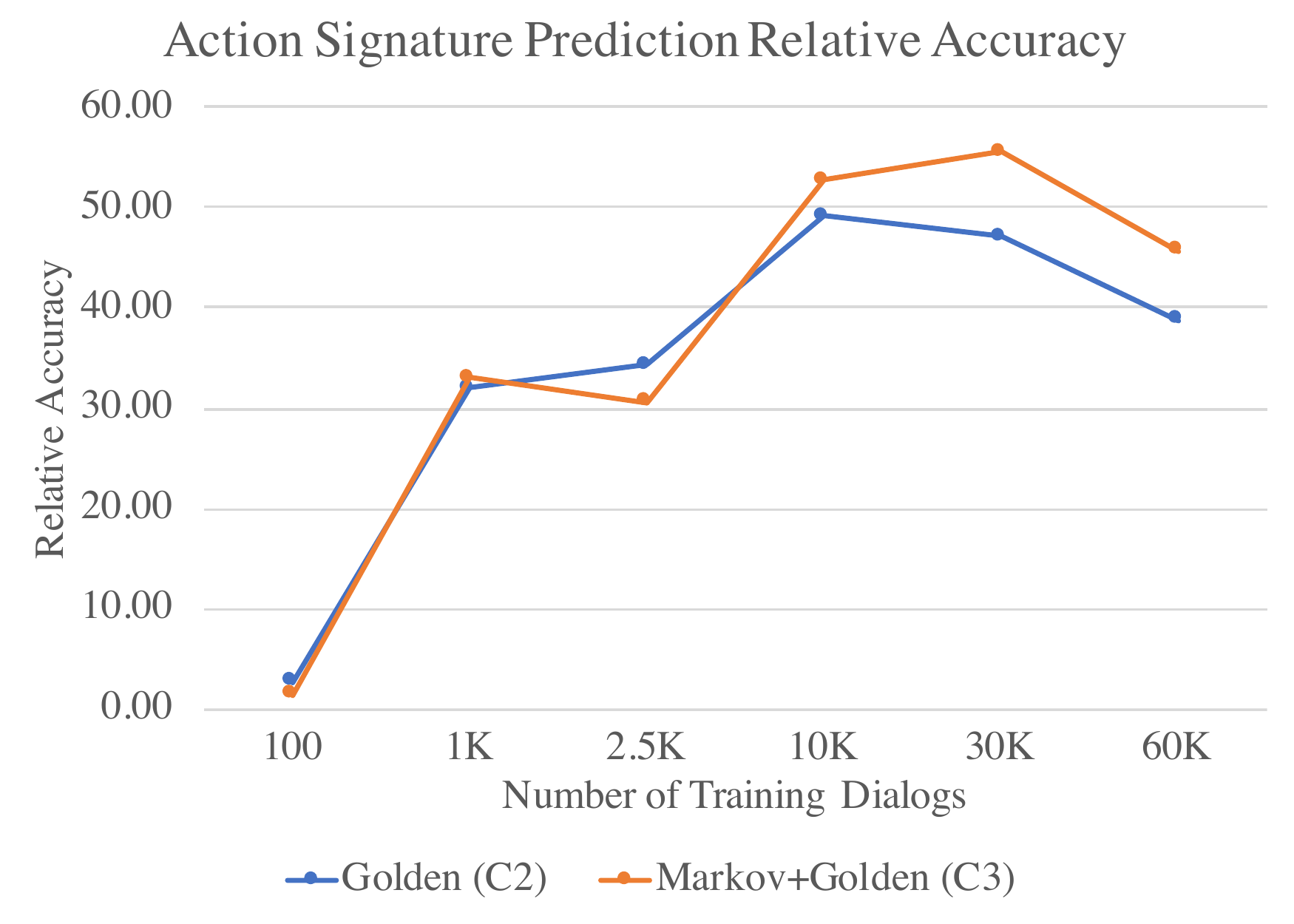}
\endminipage
\caption{NER F1 score, AP accuracy, and ASP accuracy, trained on different numbers of dialogs on the ticket booking test set, averaged 5 times}\label{fig:learning_results}
\end{figure}

We repeated the same experiment set up as Table~\ref{table:ticketbot-results} with different numbers of training dialogs. Figure~\ref{fig:learning_results} shows the learning rates of the three downstream models as the models are exposed to larger numbers of synthetic training dialogs. For all three dialog sampling configurations, performance improves sharply from $0$ to $1,000$ training examples. Under C1, the performance gains plateau at $n=1,000$, as the limited logical variation present in the seed dialogs becomes fully reflected in the synthetic data the models have already seen. However, under C2 and C3, NER and ASP performance continue to improve as the models are exposed to larger numbers of synthetic dialogs and are therefore able to more precisely capture the nuances of the additional variations. In particular, for ASP accuracy, as the number of training dialogs increases above $30k$, we observe a gap in performance between the C2 and C3 trials indicating the incremental benefit of injecting variation in \emph{user intents} by the \textit{Markov Goal Sampler}, beyond variation only in the order \& number of entity elicitations introduced by the \textit{Golden Goal Sampler}.
In this asymptotic case, both Golden and Markov continue to dramatically outperform the baseline.

\vspace{-0.2cm}
\section{Conclusion}

In this paper, we propose a novel approach to generate annotated dialogs for training goal-oriented dialog systems. Given an input schema and a few seed dialogs, the proposed approach can efficiently generate thousands of annotated dialogs with much larger dialog variation than what is present in the seed dialogs. The generated data can be used to train downstream conversational models required for goal-oriented dialog systems. As a result, application developers can build well performing downstream dialog applications without having to collect and annotate extensive training dialogs. To validate the usefulness of the proposed approach, we compare it with a baseline dialog generation strategy that randomly samples language variations and entity values from available catalogs. We evaluate both the generated dialog diversity and the downstream system performance and show that the proposed approach leads to greater dialog diversity and significantly higher downstream conversational model accuracy compared to the simpler baseline.







\bibliography{neurips_2020}{}
\bibliographystyle{plain}

\end{document}